%% file: main.tex
\documentclass[letterpaper]{article} 
\usepackage{aaai25}  
\usepackage{times}  
\usepackage{helvet}  
\usepackage{courier}  
\usepackage[hyphens]{url}  
\usepackage{graphicx} 
\usepackage{natbib}  
\usepackage{caption} 
\usepackage{algorithm}
\usepackage{algorithmic}
\usepackage{amsmath}
\usepackage{amssymb} 
\usepackage{booktabs} 
\usepackage[many]{tcolorbox}    	
\input{def}

\usepackage{newfloat}
\usepackage{listings}
\DeclareCaptionStyle{ruled}{labelfont=normalfont,labelsep=colon,strut=off} 
\floatstyle{ruled}
\newfloat{listing}{tb}{lst}{}
\floatname{listing}{Listing}

\title{SoundBrush: Sound as a Brush for Visual Scene Editing}
\author{
    Kim Sung-Bin\textsuperscript{\rm 1}, Kim Jun-Seong\textsuperscript{\rm 1}, Junseok Ko\textsuperscript{\rm 2}, Yewon Kim\textsuperscript{\rm 1}, Tae-Hyun Oh\textsuperscript{\rm 1,3,4}\\
}
\affiliations{
    \textsuperscript{\rm 1}Department of Electrical Engineering and \textsuperscript{\rm 3}Graduate School of Artificial Intelligence, POSTECH, Korea\\
    \textsuperscript{\rm 2}Department of Statistics, Inha University, Korea\\
    \textsuperscript{\rm 4}Institute for Convergence Research and Education in Advanced Technology, Yonsei University, Korea\\
}

\usepackage{bibentry}

\begin{document}

\twocolumn[{%
\renewcommand\twocolumn[1][]{#1}%
\maketitle
\input{figs/teaser}
}]

\begin{abstract}
We propose SoundBrush, a model that uses sound as a brush to edit and manipulate visual scenes. We extend the generative capabilities of the Latent Diffusion Model (LDM) to incorporate audio information for editing visual scenes. Inspired by existing image-editing works, we frame this task as a supervised learning problem and leverage various off-the-shelf models to construct a sound-paired visual scene dataset for training. This richly generated dataset enables SoundBrush to learn to map audio features into the textual space of the LDM, allowing for visual scene editing guided by diverse in-the-wild sound. Unlike existing methods, SoundBrush can accurately manipulate the overall scenery or even insert sounding objects to best match the audio inputs while preserving the original content. Furthermore, by integrating with novel view synthesis techniques, our framework can be extended to edit 3D scenes, facilitating sound-driven 3D scene manipulation. Demos are available at \url{https://soundbrush.github.io/}.
\end{abstract}

\section{Introduction}
\input{1_intro}

\section{Related Work}
\input{2_related_work}

\section{Method}
\input{3_method}

\section{Experiments}
\input{4_exp}
\section{Discussion and Conclusion}
\input{5_conclusion}

\section{Acknowledgments}
This work was supported by IITP grant funded by Korea government (MSIT) (No.RS-2023-00225630, Development of Artificial Intelligence for Text-based 3D Movie Generation; No.RS-2024-00457882, National AI Research Lab Project; No.RS-2024-00358135, Corner Vision: Learning to Look Around the Corner through Multi-modal Signals; No.RS-2024-00395401, Development of VFX creation and combination using generative AI).

\bibliography{aaai25}

\clearpage
\input{supp}

\end{document}

%% file: def.tex
\usepackage{enumitem}
\usepackage{multirow}
\usepackage{colortbl}
\usepackage{microtype}
\usepackage{comment}
\usepackage[hang,flushmargin]{footmisc}
\usepackage{graphicx}
\usepackage{multirow}
\usepackage[font=small,labelfont=bf]{caption}

\setlist[itemize]{align=parleft,left=0pt,topsep=1mm,itemsep=0mm,parsep=1mm}
\definecolor{azure}{rgb}{0.0, 0.5, 1.0}
\definecolor{azure(colorwheel)}{rgb}{0.0, 0.5, 1.0}
\definecolor{nicegreen}{rgb}{0.0, 0.7, 0.1}
\definecolor{yw}{rgb}{0.01176, 0.5490, 0.5490}
\definecolor{ashblue}{rgb}{0.36, 0.54, 0.66}
\definecolor{ashgrey}{rgb}{0.7, 0.75, 0.71}
\definecolor{applegreen}{rgb}{0.55, 0.71, 0.0}
\definecolor{blue}{rgb}{0.0, 0.0, 1.0}
\definecolor{postechred}{rgb}{0.784, 0.003, 0.313}
\definecolor{ywg}{rgb}{0.9960, 0.8984, 0.5859}
\definecolor{ballblue}{rgb}{0.13, 0.67, 0.8}
\definecolor{cornellred}{rgb}{0.7, 0.11, 0.11}
\definecolor{darkcyan}{rgb}{0.0, 0.55, 0.55}
\definecolor{CuGray}{gray}{0.9}
\definecolor{airforceblue}{rgb}{0.36, 0.54, 0.66}
\definecolor{rev}{rgb}{0.784, 0.003, 0.313}
\definecolor{pink}{cmyk}{0, 0.7808, 0.4429, 0.1412}
\definecolor{amethyst}{rgb}{0.6, 0.4, 0.8}
\definecolor{black}{rgb}{0.0, 0.0, 0.0}
\definecolor{tb3_yellow}{rgb}{0.996, 1.0, 0.6}
\definecolor{tb3_orange}{rgb}{0.980, 0.8, 0.604}
\definecolor{tb3_red}{rgb}{0.972, 0.6, 0.6}
\definecolor{dimgray}{rgb}{0.41, 0.41, 0.41}
\definecolor{brickred}{rgb}{0.8, 0.25, 0.33}
\definecolor{bleudefrance}{rgb}{0.19, 0.55, 0.91}
\definecolor{blue(ncs)}{rgb}{0.265, 0.445, 0.765}
\definecolor{blue(ryb)}{rgb}{0.01, 0.28, 1.0}
\definecolor{orange}{rgb}{1.0, 0.49, 0.0}
\definecolor{Gray}{gray}{0.88}
\definecolor{green(ncs)}{rgb}{0.0, 0.62, 0.42}
\definecolor{clova}{rgb}{0.24, 0.63, 0.33}
\definecolor{correctgreen}{rgb}{0.01, 0.702, 0.321}
\definecolor{wrongred}{rgb}{0.757, 0, 0}

\definecolor{teaserblue}{rgb}{0.274, 0.694, 1}
\definecolor{teaserorange}{rgb}{0.913, 0.443, 0.196}
\definecolor{teaserred}{rgb}{1,0,0}

\definecolor{kellygreen}{rgb}{0.3, 0.73, 0.09}

\newcolumntype{g}{>{\columncolor{CuGray}}c}
\newcolumntype{z}{>{\columncolor{CuGray}}l}

\renewcommand{\paragraph}[1]{\noindent\textbf{#1.}\,\,}

\definecolor{steelblue}{rgb}{0.27, 0.51, 0.7}

\usepackage{xspace}

\makeatletter
\def\@fnsymbol#1{\ensuremath{\ifcase#1\or *\or \dagger\or \ddagger\or
   \mathsection\or \mathparagraph\or \|\or **\or \dagger\dagger
   \or \ddagger\ddagger \else\@ctrerr\fi}}
\makeatother

\def\onedot{.\@\xspace}
\def\eg{\emph{e.g}\onedot} 
\def\ie{\emph{i.e}\onedot}

\newcommand{\Eref}[1]{Eq.~(\ref{#1})}
\newcommand{\Fref}[1]{Fig.~\ref{#1}}
\newcommand{\Tref}[1]{Table~\ref{#1}}

\newcommand{\be}{\begin{eqnarray}}
\newcommand{\ee}{\end{eqnarray}}
\newcommand{\bee}{\begin{eqnarray*}}
\newcommand{\eee}{\end{eqnarray*}}

\newcommand{\matrixb}{\left[ \begin{array}}
\newcommand{\matrixe}{\end{array} \right]}

\usepackage{mathtools}
\DeclarePairedDelimiter{\norm}{\lVert}{\rVert}

\newtcolorbox{boxA}{
    boxrule = 0.8pt,
    colframe = black 
}

%% file: figs/teaser.tex
\vspace{-8.5mm}
\captionsetup{type=figure}
\includegraphics[width=1.0\linewidth]{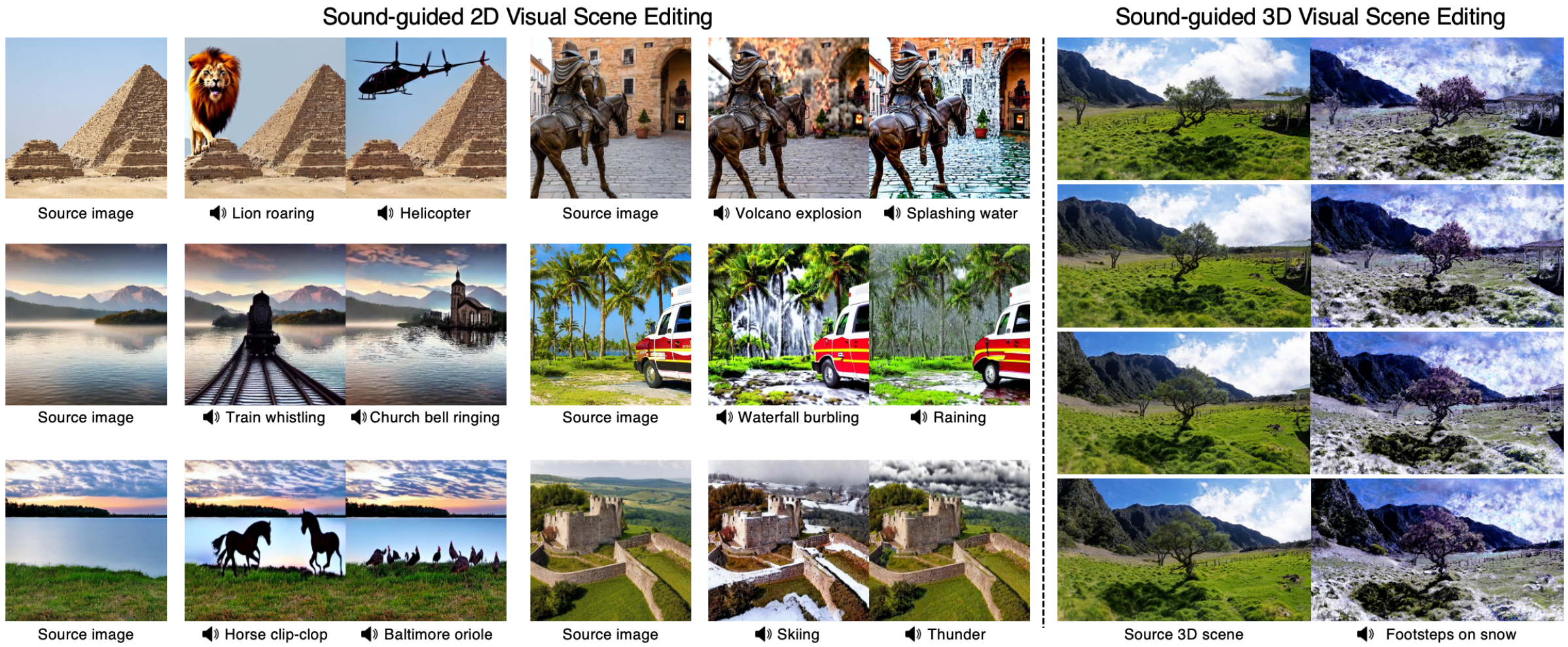}
\vspace{-7mm}
\captionof{figure}{{\bf Sound-guided visual scene editing.} We propose SoundBrush, a model that can modify visual scenes by adding sounding objects and adjusting the scene to align with the input sound (left). Furthermore, it can be extended to edit 3D visual scenes using the input sound (right).}
\vspace{4mm}
\label{fig:teaser}

%% file: 1_intro.tex
Visual scene editing extends beyond image generation by modifying existing images to meet specific user requirements. With its promising applications, this field has evolved to include tasks, such as style transfer~\cite{gatys2016image} and translation between image domains~\cite{zhu2017unpaired,huang2018multimodal}. Recently, the remarkable capabilities of text-based generative models have significantly broadened the scope of visual scene editing to include not only 2D images~\cite{pnp, p2p, pix2pix} but also 3D scenes~\cite{n2n}, guided by text descriptions as control signals.

Although text descriptions are effective, sound has become a unique control signal for visual scene editing~\cite{lee2022sound, avstyle, any2pix}. Unlike text, sound shares the natural association with visual scenes and can provide rich information that text may overlook. For example, the intensity of rainfall can range from light to heavy in sound, whereas a text description might simply note it as ``raining.'' Despite these advantages, existing sound-guided editing methods face significant challenges, such as being limited to generating entirely new scenes rather than editing existing ones~\cite{sound2scene, audiotoken, gluegen, sung2024sound2vision}, struggling to insert new objects while only manipulating styles~\cite{lee2022sound, avstyle}, and altering the structure of the source image after editing~\cite{any2pix}.

Addressing these limitations requires overcoming several challenges. First, effectively capturing informative audio signals to determine how to manipulate or edit the given image is difficult due to the significant information gap between audio and visual signals. Secondly, the scarcity of sound-paired image editing datasets and the difficulty in acquiring them further amplify these challenges, becoming a bottleneck for learning the capabilities of sound-guided visual scene editing.

In this work, we propose SoundBrush, a model that tackles these challenges and uses sound as a brush to edit visual scenes. Specifically, we extend the generative capabilities of the text-to-image model, Latent Diffusion Model (LDM)~\cite{ldm}, by incorporating audio information into visual scene editing. 
To utilize sound as an editing control signal, we augment textual token spaces of LDM to include diverse auditory features and design a mapping network that converts sound into these tokens. 
Inspired by existing image-editing works~\cite{pix2pix}, we employ various off-the-shelf models, such as sound source localization~\cite{park2024can}, image inpainting~\cite{suvorov2022resolution}, and Prompt-to-Prompt model~\cite{p2p}, to generate a sound-paired visual scene editing dataset for training. This richly generated dataset facilitates joint training of the mapping network and LDM in SoundBrush, enabling the model to edit visual scenes guided by various in-the-wild sound cues.

We validate the efficacy of our proposed SoundBrush by comparing it with existing sound-guided visual scene editing models~\cite{audiotoken,gluegen,any2pix}. Unlike previous methods, SoundBrush can accurately insert sounding objects and edit the overall scenery to reflect the sound semantics, as shown in \Fref{fig:teaser}. Furthermore, by integrating with a novel view synthesis method~\cite{nerf}, our framework can be extended to edit 3D scenes, enabling sound-guided 3D scene editing. Our main contributions are summarized as follows:

\begin{itemize}
\item Proposing SoundBrush, a model that effectively incorporates auditory information to manipulate visual scenes.
\item Generating a comprehensive dataset paired with sound cues and visual data, which facilitates the training of models for sound-guided visual scene editing.
\item Demonstrating SoundBrush's ability to accurately insert objects or manipulate the overall visual scenes based on sound cues, including extensions to 3D scene editing.
\end{itemize}

%% file: 2_related_work.tex
\paragraph{Multimodal-guided image editing}
Image editing aims to modify source images according to diverse user requirements. Recent advancements, including image-to-image translation~\cite{isola2017image, liu2017unsupervised, huang2018multimodal}, object insertion~\cite{any2pix,audiotoken}, and 3D scene editing~\cite{n2n}, have leveraged various modalities to meet diverse needs of applications. Driven by the generative power and flexibility of Latent Diffusion Models (LDMs)~\cite{ldm}, integrating multiple modalities within LDMs has become foundational for manipulating images using different user inputs, such as images~\cite{paint}, drag-and-click actions~\cite{drag}, and textual descriptions~\cite{edit}.

Among these modalities, textual description is a predominant signal for editing, allowing users to describe their editing intentions using words. The introduction of the CLIP model~\cite{clip} has significantly advanced this area by leveraging high-level text-visual embeddings to guide image editing. However, achieving precise control over image changes and ensuring consistent results for specific prompts remain challenging. To address these challenges, methods such as Prompt-to-Prompt~\cite{p2p} and Plug-and-Play~\cite{pnp} have shown that controlling the attention map within a diffusion model—originally trained for image generation—can effectively edit images without additional training. These approaches enhance the precision and consistency of edits based on user inputs. 
Building on the success of text-based methods~\cite{pix2pix, imagen, diffusionclip}, the expansion to other input modalities continues to progress by aligning each modality to these text-based models~\cite{any2pix, Imagic}. In this trend, our main goal is to extend the existing text-based editing model to include in-the-wild sound as a control modality. By leveraging expressive semantics in audio inputs, we aim to significantly advance multimodal image editing.

\paragraph{Sound-guided image synthesis}
Recent studies in sound-guided image synthesis have shown that the cross-modal association between sight and sound provides important information effectively leveraged for image synthesis. Utilizing this fact, one line of research focuses on generating images from sound inputs. Initially, early works targeted specific sound categories, such as musical instruments~\cite{cmcgan, strummingbeat}, bird sounds~\cite{s2b}, or speech data~\cite{speech2face}, for image synthesis. With advancements in generative models, more recent methods have achieved remarkable results using Generative Adversarial Networks (GANs)~\cite{sound2scene} or text-conditioned diffusion models~\cite{audiotoken, gluegen}. However, since their main goal is generation rather than editing, these methods often struggle to preserve the original content when used to edit images with sound.

In response to this, another line of research has emerged that proposes editing existing images using sound-based inputs. \citet{lee2022sound} expand the embedding space of text-based image manipulation models to include sound inputs, and \citet{avstyle} utilize conditional GANs to adjust the visual style of images to match sounds by learning from unlabeled audio-visual data. More recently, InstructAny2Pix~\cite{any2pix} shows significant improvement in image editing by introducing a model that takes multimodal instructional inputs, including sound and text. Despite their success, these existing sound-guided image editing models primarily focus on style manipulation rather than the insertion of sounding objects~\cite{lee2022sound, avstyle} and often fail to preserve the content of the source image~\cite{any2pix}. Our work aligns with the latter approach, proposing a model that can edit and manipulate visual scenes based on sound inputs. Furthermore, we aim to address the limitations of existing methods by accurately inserting sounding objects and editing scenes based on sound, including extensions to 3D scene editing~\cite{n2n}.

\input{figs/data_pipeline}

%% file: figs/data_pipeline.tex
\begin{figure*}[tp]
    \centering
    \small
    \includegraphics[width=1.0\linewidth]{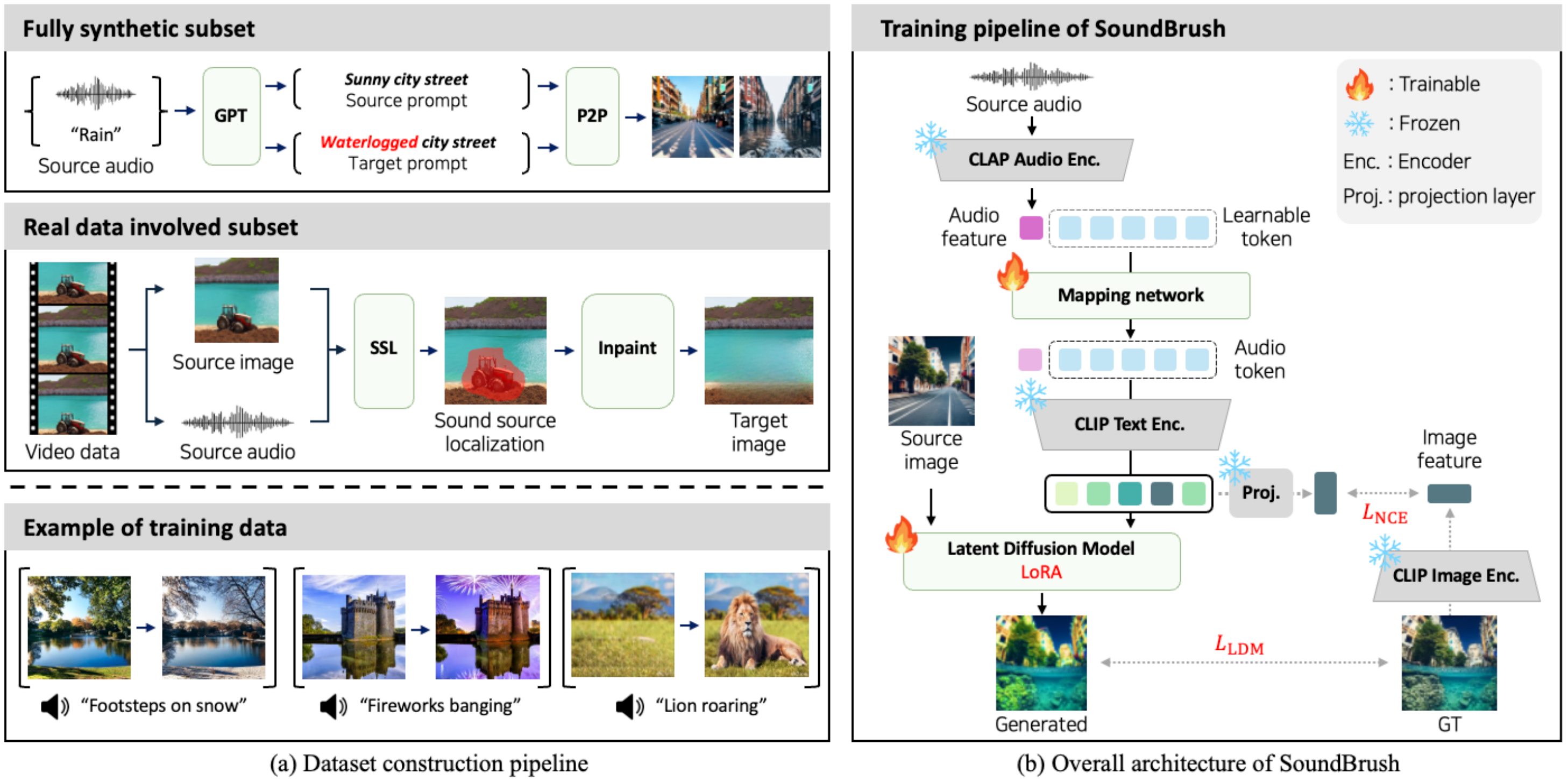}
    \vspace{-5mm}
    \caption{\textbf{Our proposed approach.} We start by designing an automatic dataset-construction pipeline as in (a). The dataset is constructed with a fully synthetic subset, involving synthetically generated image pairs paired with audio, and real data involved subset, involving real audio and images. Using this dataset, we train SoundBrush to effectively translate audio features into audio tokens, which can then be used as control signals for the image editing latent diffusion model as described in (b).}
    \label{fig:pipeline}
\end{figure*}

%% file: 3_method.tex
Our goal in this work is to edit visual scenes based on the given in-the-wild audio. We are motivated by the fact that sound contains useful information that can directly influence the editing of an image. This information in the sound may be related to a sound-producing object, \eg, a ``Dog barking'' and a ``Train whistling,'' or it could indicate general changes in the scene, \eg, ``Raining'' and ``Stream burbling.'' Inspired by the prior arts~\cite{pix2pix, any2pix}, we approach this as a supervised learning problem. We begin by generating a paired training dataset that includes audio along with images both before and after editing. This dataset is then used to train a conditional diffusion model that uses audio as a control signal, enabling to learn how to accurately edit the visual scene based on the perceived sound. The overall pipeline of training data generation is described in~\Fref{fig:pipeline} (a).

\subsection{Constructing the Training Dataset}
Leveraging synthetic datasets has proven effective for training in generative tasks \cite{pix2pix, any2pix}. Given that our dataset requires the incorporation of sound, we have extended these prior works to construct a rich dataset that pairs a sound with an input source image of a visual scene and a target image that reflects auditory information. Specifically, we develop two subsets: one fully synthetic, consisting of generated image pairs with corresponding audio, and the other synthetically generated but incorporating real images and audio.

\paragraph{Fully synthetic subset}
The generation of the fully synthetic subset involves creating source and target prompts based on sound events, followed by generating paired images from these prompts (\Fref{fig:pipeline} (a)-row 1). For example, given the source prompt ``Sunny city street'' and the sound-related keyword ``Waterlogged,'' we create the target prompt ``Waterlogged city street,'' which is then used to generate paired images. 
For this, we design an automated pipeline using GPT-4~\cite{gpt4} and a text-to-image diffusion model~\cite{ldm}. This process includes (1) selecting categories from the sound dataset, (2) creating source and target prompts with sound-related keywords, and (3) generating the corresponding before and after images.

Initially, we extract sound categories from the VGGSound dataset~\cite{vggsound}, focusing on environmental sounds like ``Waterfall burbling'' and ``Thunder.'' We then generate source prompts by combining various locations, objects, weather conditions, and environments, utilizing GPT-4’s extensive language capabilities. Subsequently, we create source-target prompts in large quantities through few-shot prompting with GPT-4, integrating sound-related keywords into the target prompts. This allows for a natural integration of various sound events into the target prompts, beyond simply concatenating keywords. The constructed prompt pairs are used for generating image pairs. Following previous studies~\cite{pix2pix}, we employ Prompt-to-Prompt~\cite{p2p} to create before and after editing samples. This method adjusts the cross-attention in the denoising process to enhance content similarity while reflecting changes specified in the target prompt.

After generating the samples, we adopt several CLIP-based metrics, including directional similarity in CLIP space~\cite{gal2022stylegan} and feature similarity between two images in CLIP space, to ensure the quality of the generated subset. Images that fall below a specified threshold are filtered out. Additionally, we incorporate ImageBind~\cite{girdhar2023imagebind} to validate whether the target image accurately represents the sound event by measuring the audio-visual feature similarity in the ImageBind space and excluding those that do not meet the required threshold.

\paragraph{Real data involved subset}
Along with the fully synthetic one, we also construct a subset that incorporates real-world data (\Fref{fig:pipeline} (a)-row 2). We begin by extracting paired audio and images from the VGGSound dataset. We then use a sound source localization model~\cite{park2024can} to detect and localize the sounding object with a coarse mask. Given this coarse mask on the sounding object, we employ LaMa~\cite{suvorov2022resolution} to inpaint over the mask of the sounding object. The inpainted image serves as the ``before editing'' image, while the original image is used as the ``after editing'' image, thus creating triplet pairs for the dataset. However, as neither the sound source localization model nor the inpainting model may be sufficiently reliable, we use ImageBind~\cite{girdhar2023imagebind} to filter out noisy inpainted pairs. We remove pairs where the feature similarity between the audio features and the inpainted image in the ImageBind space is above the threshold, \ie, indicating that the inpainted image still consists the sounding object.

In total, we construct a dataset consisting of 83,614 pairs, with 27,056 fully synthetic and 56,558 involving real data. The example pairs are shown in the \Fref{fig:pipeline} (a)-row 3.

\subsection{Learning to Edit Visual Scene with Sound}
We leverage the strong generative capabilities of the existing Latent Diffusion Model (LDM)~\cite{ldm} and extend this to accommodate sound as a condition for editing. To achieve this, we design a mapping network that translates audio features into a sequence of tokens within the textual spaces of LDM. These tokens are then fed into the LDM to manipulate the visual scene. We train this model with our constructed dataset to jointly learn how to map audio into sequences of tokens in the textual space, while also learning to edit visual scenes using these tokens. The overall pipeline of our proposed SoundBrush is illustrated in~\Fref{fig:pipeline} (b).

\paragraph{Preliminary of the text-guided image editing model}
We base our model on InstructPix2Pix~\cite{pix2pix}, which utilizes text instructions for image editing. InstructPix2Pix is built on the LDM~\cite{ldm}, which learns the underlying probabilistic model of image data $x$ within the latent space $p(z)$ of the variational autoencoder with a encoder $\mathcal{E}(\cdot)$ and a decoder $\mathcal{D}(\cdot)$, where $z = \mathcal{E}(x)$, and $\hat{x} = \mathcal{D}(\mathcal{E}(x))$.
Learning such probabilistic model involves learning the reverse Markov process over a sequence of $T$ timesteps in the latent space. 
For each timestep $t=0,\dotsc,T$, the denoising function $\epsilon_{\theta}: R^d \rightarrow R^d$, where $d$ is the dimension of the latent, is trained to predict a denoised version of the perturbed $z_t$ at timesetp $t$, as $\epsilon_{\theta}(z_t,t)$.

InstructPix2Pix incorporates the conditioning text input $c^V$ during the denoising process by employing the CLIP text encoder as $c^V=\text{CLIP}_\text{T}{(V)}$, where $V$ represents the textual tokens of the input text instruction.
Additionally, InstructPix2Pix adds extra input channels to the first convolutional layer of the LDM to facilitate image conditioning by concatenating
$z_t$ with $\mathcal{E}(c^I)$, where $c^I$ is the conditioning image.
The corresponding objective can be simplify written as follows: 
\begin{equation}\label{eq1}
L_{\text{LDM}}=\mathbb{E}_{z_t, t, \epsilon\in \mathcal{N}(0,I)}[\norm{\epsilon-\epsilon_\theta(z_t, t, \mathcal{E}(c^I), c^V)}_2^2]. 
\end{equation}

\paragraph{Translating audio into textual spaces}
While the conditional input $c^V$ in InstructPix2Pix is derived from text instructions, our aim is to use sound solely as the conditional input to guide the model in editing and manipulating visual scenes.
To facilitate this, we design a mapping network that translates audio features into tokens within the textual spaces of the LDM. 
Specifically, given the input audio $A$, we extract audio features $f^A=\text{CLAP}(A)$, where $\text{CLAP}(\cdot)$ is a pretrained CLAP audio encoder~\cite{wu2023large}.
These features $f^A$ are then fed into the mapping network $M(\cdot)$, which converts $f^A$ into a sequence of audio tokens in the textual spaces as $V^A=M(f^A)$. 
The mapping network, built with Transformer encoder layers, takes two different inputs: the audio features and a sequence of learnable tokens. 
These learnable tokens are designed to extract useful information from the audio features through multi-head attention, while adjusting the tokens to formulate a signal that enables the model to perform editing. Finally, we feed the sequence of
audio tokens
into the CLIP text encoder to extract text-aligned audio conditions as ${c^A}=\text{CLIP}_\text{T}(V^A)$. The extracted $c^A$ can then replace $c^V$ for sound-guided visual scene editing.

\input{figs/2d_qual}

\paragraph{Learning objectives}
One straightforward way to train the mapping network is by optimizing \Eref{eq1} with $c^V$ replaced by $c^A$. However, we find that soley optimizing \Eref{eq1} is insufficient for the mapping network to learn to translate audio into semantically meaningful tokens. 
To extract useful information from the audio features and effectively map them into textual space in a continuous form, we propose incorporating a contrastive loss between the features derived from audio tokens and the visual features of the ground-truth image. 
The audio conditions ${c^A}$
are fed into the projection layer of the CLIP text encoder to extract features, $q^V=P(c^A)$, where $P(\cdot)$ is the projection layer. The ground-truth image $I$ is fed into the CLIP visual encoder to extract visual features, $q^I=\text{CLIP}_\text{I}(I)$. We employ the InfoNCE loss~\cite{oord2018representation}, treating pairs of $q^V$ and $q^I$ as positive and those from different pairs as negative in the batch $N$. The objective is formulated as:
\begin{equation}\label{eq2}
L_{\text{NCE}} = \frac{1}{N}\sum^N_{j=1}-\log{\tfrac{\exp(\langle q^V_j, q^I_j\rangle)}{\sum^N_{k=1}\exp(\langle q^V_j, q^I_k\rangle)}}, 
\end{equation}
where $\langle\cdot\rangle$ is the cosine similarity function. Furthermore, although InstructPix2Pix has demonstrated rich editing capabilities, we find that the model struggles to insert new objects. Therefore, we jointly optimize the mapping network and LDM using Low-Rank Adaptation (LoRA)~\cite{lora}, while the CLIP text encoder and the CLAP audio encoder remain fixed. 
Finally, we add an $\ell_1$ regularization to the audio tokens to encourage a more even distribution. 
Thus, our final objective for training SoundBrush is as follows:
\begin{equation}
L_{\text{total}} = L_{\text{LDM}}+\lambda_{\text{NCE}} L_{\text{NCE}}+\lambda_{\ell_1} |V^A|_1.
\end{equation}

%% file: figs/2d_qual.tex
\begin{figure*}[tp]
    \centering
    \small
    \includegraphics[width=1.0\linewidth]{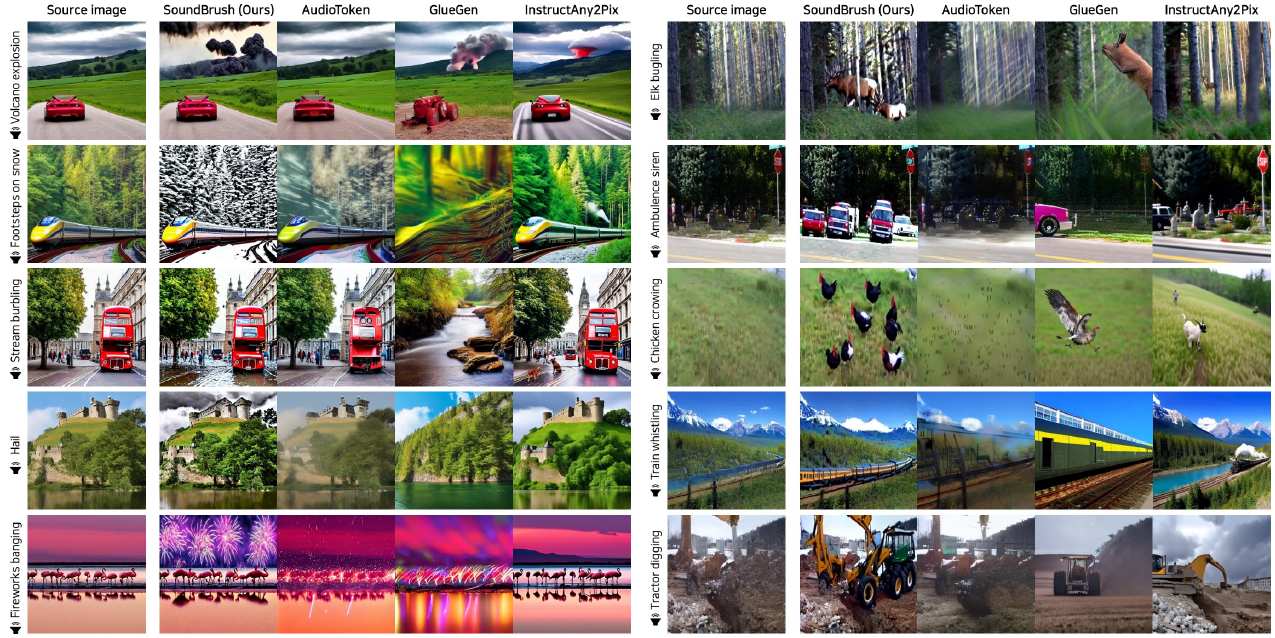}
    \caption{\textbf{Qualitative comparison.} We compare our model with existing sound-guided visual scene editing methods and demonstrate that our model can edit visual scenes using diverse in-the-wild audio while preserving the rest of the content unchanged.}
    \label{fig:2d_qual}
\end{figure*}

%% file: 4_exp.tex
We validate the editing power of our proposed SoundBrush both qualitatively and quantitatively. We begin by outlining the experimental setup, which includes the dataset, metrics, and competing methods. We then present comparisons of sound-guided 2D visual scene editing between SoundBrush and existing methods. Finally, we demonstrate how SoundBrush can be extended to edit 3D visual scenes.

\subsection{Experimental Setup}\label{exp:setup}
\paragraph{Dataset}
We construct the evaluation dataset following the previously described dataset construction pipeline. All the audio files are sourced from VGGSound~\cite{vggsound}, which is an audio-visual dataset containing around 200K videos from 309 different sound categories. We select 20 categories from these and use the provided test splits for constructing the evaluation dataset.

\paragraph{Evaluation metrics}
We assess our method using both objective and subjective metrics. For objective metrics, we measure audio-visual similarity (AVS) using ImageBind~\cite{girdhar2023imagebind} space by computing the feature similarity between the input audio and the edited image. 
We also evaluate image-image similarity (IIS) in the CLIP~\cite{clip} space by comparing the feature similarity between edited and ground-truth images. 
Text-visual similarity (TVS) is assessed in the CLIP space, evaluating the similarity between the audio category's name and the edited image. 
Additionally, we measure the Fr\'{e}chet Inception Distance (FID), which quantifies the distance between the features obtained from real and synthesized images using a pre-trained Inception-V3~\cite{inceptionv3}.
For subjective metrics, we conduct human evaluations to analyze performance from a human perception perspective, focusing on evaluating whether the edited images remain similar to the original images while accurately reflecting the conditioned sound.

\paragraph{Competing methods}
We compare the editing capabilities of our method against three different methods: AudioToken~\cite{audiotoken}, GlueGen~\cite{gluegen}, and InstructAny2Pix~\cite{any2pix}. 
AudioToken and GlueGen are originally targeted for generating images from sound and have demonstrated significant image generation performance. 
To adapt these models for image editing, we employ a training-free Plug-and-Play (PnP) method~\cite{pnp} that enables them to edit visual scenes with sound input. Additionally, as GlueGen is initially trained with UrbanSound8K~\cite{urban}, we fine-tune this model using the VGGSound dataset to ensure a fair comparison. 
We also benchmark against InstructAny2Pix, which utilizes sound and instructional text for image editing.

\subsection{Results on 2D Image Editing}
\paragraph{Qualitative results}
Figure~\ref{fig:2d_qual} shows a qualitative comparison between existing methods~\cite{audiotoken, gluegen, any2pix} and our proposed SoundBrush. 
As demonstrated, our model exhibits superior capability in manipulating and editing visual scenes based on audio inputs. AudioToken struggles to reflect the audio signal in the given image while preserving the original structure, and GlueGen reflects the audio signal but modifies the original content beyond recognition.
InstructAny2Pix shows favorable editing results; however, if we take a closer look, the original content has been altered.
For instance, InstructAny2Pix changes the design of the car and the road when editing with the ``Volcano explosion'' sound, as shown in the first row of \Fref{fig:2d_qual}. The ``Tractor digging'' sound in the bottom right sample successfully inserts the tractor but also significantly changes the structure of the original image. 
In contrast, our proposed SoundBrush successfully manipulates the overall scenery or inserts a sounding object into the scene, all while maintaining the original structure of the given image.

\begin{figure}[tp]
    \centering
    \small
    
    \includegraphics[width=1.0\linewidth]{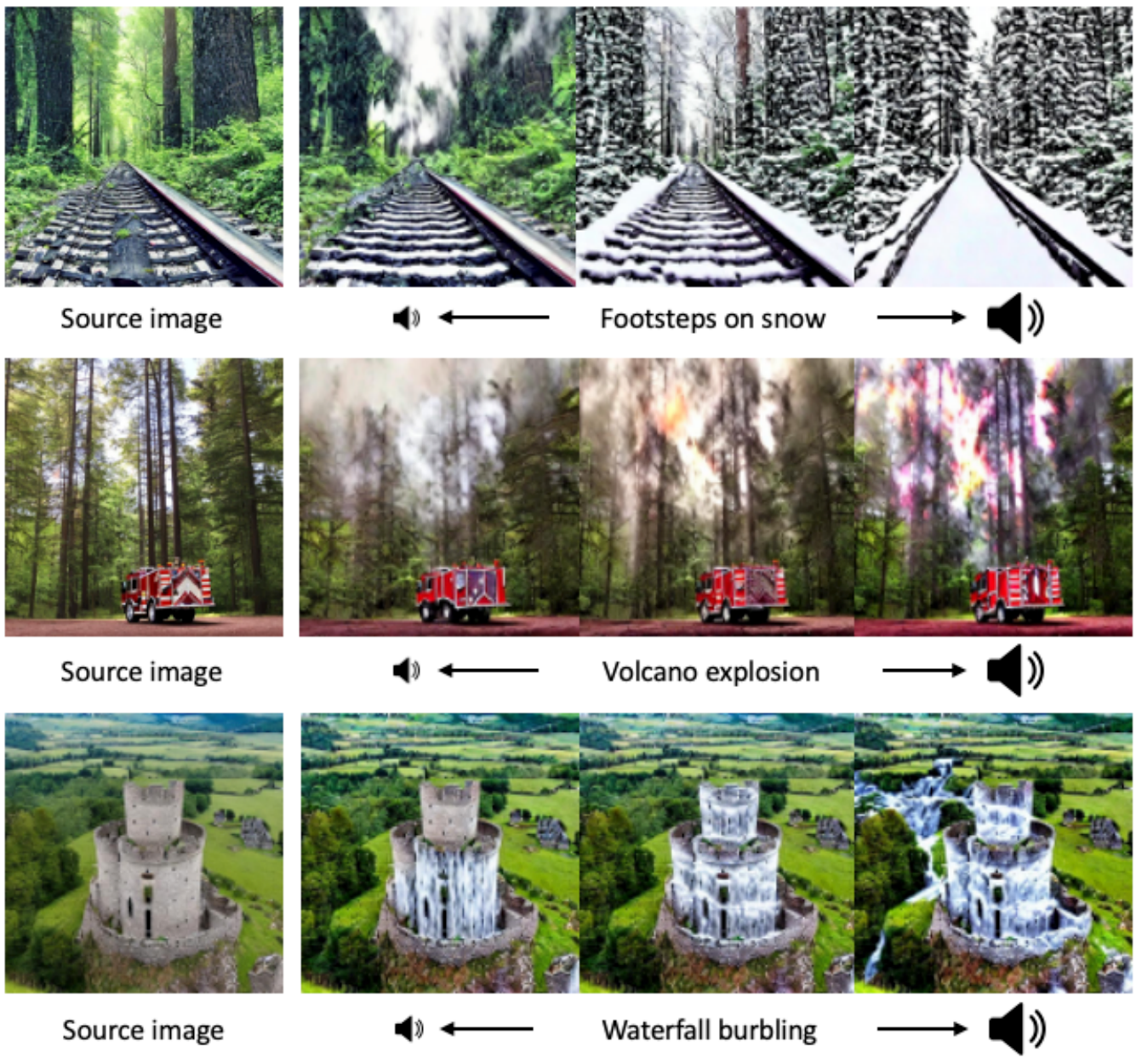}
    \caption{\textbf{Editing results by different audio intensities.} SoundBrush captures the intensity differences in the audio and reflects these changes in the edited images.}
    \label{fig:qual_v2}
\end{figure}

In addition, although not explicitly modeled, we observe that our model can reflect the intensity variations inherent in sound during editing, as shown in \Fref{fig:qual_v2}. This is a distinct property compared to text, where sound uniquely carries intensity information that can be utilized for editing images. We assess this by manipulating the volume of the reference audio and feeding these varied-volume audios into the model for visual scene editing. For example, as the ``Footsteps on snow'' sound becomes larger, the railroad begins to be covered with denser snow; similarly, as the ``Volcano explosion'' sound increases, the fire and smoke become larger and more dynamic. These results support that our model not only has a category-specific understanding of audio but also perceives the relationship between audio volume and visual changes.

\begin{figure}[t]
    \small
    \centering
    \resizebox{0.95\linewidth}{!}{
    \begin{tabular}{lcccc}
    \toprule
     Method & AVS ($\uparrow$) & IIS ($\uparrow$) & TVS ($\uparrow$) & FID ($\downarrow$)\\
    \cmidrule{1-5}
    AudioToken & 0.204& 0.663& 0.168 & 133.3\\
    GlueGen & 0.238& 0.628& 0.202 & 214.2\\
    InstructAny2Pix & 0.222& 0.729& 0.172 & \textbf{126.5}\\
    SoundBrush (Ours) & \textbf{0.261}& \textbf{0.772}& \textbf{0.203} & 131.7\\
    \bottomrule\\
    \end{tabular}
    }
    \\ \vspace{-2mm}\scriptsize (a) Quantitative comaprison\\[2mm]
    
    \includegraphics[width=0.95\linewidth]{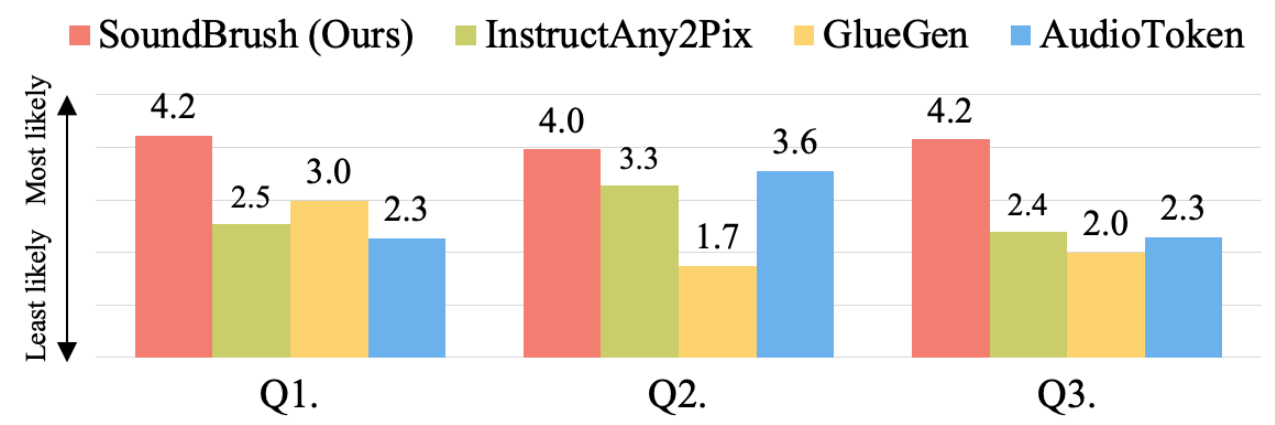} \\
    \vspace{-2mm} (b) Human evaluations
    \caption{\textbf{Quantitative comparison and user study.} We compare SoundBrush with existing methods and demonstrate that it outperforms them in overall metrics (a). We also provide human evaluation results that align with the quantitative comparisons (b).}
    \label{fig:quan}
    
\end{figure}

\paragraph{Quantitative results}
We present quantitative results to further analyze model comparisons. As summarized in \Fref{fig:quan} (a), our model consistently outperforms other methods in overall metrics, particularly excelling in the audio-visual similarity (AVS) and image-image similarity (IIS) metrics. While InstructAny2Pix achieves better performance on the FID, it shows degradation in other metrics, indicating that it can generate realistic images but often fails to preserve the original content or accurately reflect the sound semantics in the edited images. These results highlight our model's ability to reflect the audio condition effectively while maintaining the original content and structure of the given image.

\paragraph{User study}
We analyze the performance of our proposed SoundBrush from a human perception perspective. We evaluate it based on three criteria, each scored from 1 to 5: Q1. how well the input audio is reflected in the edited image, Q2. how well the original structure and content remain unchanged, and Q3. how well the audio information is reflected while original structure unchanged. 
To conduct this evaluation, we recruit 47 participants to respond to questions about 20 different randomly ordered samples.
Figure~\ref{fig:quan} (b) summarizes the results. As shown, our model is preferred by the humans in all the criterias. Especially Q3, which entails both of Q1 and Q2, our model significantly outperforms the other methods which is aligned with the quantitative evaluations.

\paragraph{Ablation study}
We conduct a series of experiments to verify our design choices as detailed in \Tref{tab:ablation}. We evaluate the impact of varying the number of audio tokens in the mapping network and the effect of applying the loss function specified in \Eref{eq2}. 
We find that using a single audio token (A) does not contain sufficient information for sound-guided image editing.
Increasing the number of tokens to five (B) results in significant improvements; however, further increasing to ten (C) begins to degrade performance.
Additionally, we validate \Eref{eq2} by comparing results with configurations (B) and (C) and observe that applying \Eref{eq2} stabilizes model training and leads to improved performance. These insights have guided us to our final model configuration (C).

\begin{table}[t]
\footnotesize
\centering
    \resizebox{1\linewidth}{!}{
    \begin{tabular}{l@{\quad}c@{\quad}c@{\quad}c@{\quad}c@{\quad}c@{\quad}c}
    \toprule
    &Num. of tokens&$L_{\text{NCE}}$&AVS ($\uparrow$) & IIS ($\uparrow$) & TVS ($\uparrow$) & FID ($\downarrow$)\\ 
    \cmidrule{1-7}
    
    (A) &1 & \checkmark&0.235&0.761&0.192 & 131.8\\
    (B) & 5& - &0.258&0.767&0.199 & 138.3\\
    (C) & 5& \checkmark&\textbf{0.261}& \textbf{0.772}&\textbf{0.203} & \textbf{131.7}\\
    (D) &10&\checkmark&0.233&0.763&0.191 & 136.1\\
    \bottomrule
    \end{tabular}
    }
    \caption{\textbf{Ablation studies of our proposed method.} We evaluate the effectiveness of various design choices by comparing different configurations of our method, varying the number of audio tokens and adapting the $L_{\text{NCE}}$ loss function during model training.}
\label{tab:ablation}
\end{table}

\input{figs/3d_qual}

\subsection{Results on 3D Image Editing}\label{exp:3d}
To further validate sound-guided image editing performance, we conduct a novel experiment called sound-based 3D scene editing. This experiment utilizes the InstructNeRF2NeRF pipeline~\cite{n2n}, which progressively edits and updates training images using InstructPix2Pix as learning progresses. For this experiment, we modify the InstructNeRF2NeRF model by replacing InstructPix2Pix with either InstructAny2Pix or our proposed SoundBrush.

Effective 3D scene editing requires making changes only where necessary, without altering the overall structure or content of the image. As illustrated in \Fref{fig:3d_qual}, InstructAny2Pix leads to inconsistencies in editing across multi-view images, resulting in floating and blurry artifacts in 3D editing. In contrast, our model tends to preserve the original structure while making appropriate changes, achieving accurate 3D editing results with coherent object occupancy.
These results demonstrate our model's potential to be further extended for editing 3D visual scenes based on sound cues. 

%% file: figs/3d_qual.tex
\begin{figure*}[tp]
    \centering
    \small
    \includegraphics[width=1.0\linewidth]{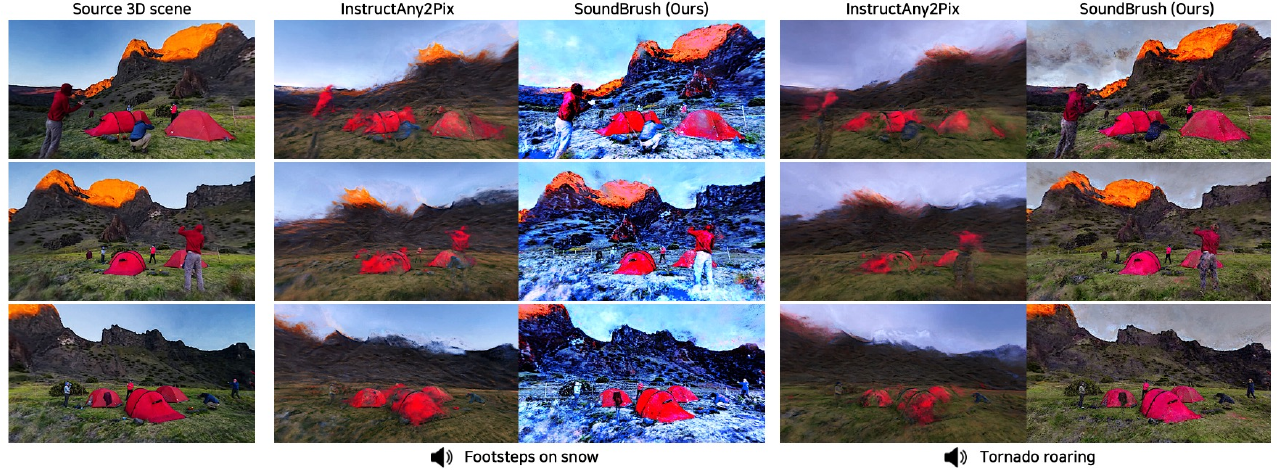}
    \vspace{-3mm}
    \caption{\textbf{3D visual scene editing results}. We compare SoundBrush with InstructAny2Pix in 3D scene editing. SoundBrush effectively edits the input image while preserving the original geometric properties, achieving successful editing in 3D scenes. In contrast, InstructAny2Pix struggles to retain the original content and thus fails to accurately capture the 3D structure.}
    
    \label{fig:3d_qual}
\end{figure*}

%% file: 5_conclusion.tex
\begin{figure}[tp]
    \centering
    \small
    \includegraphics[width=1.0\linewidth]{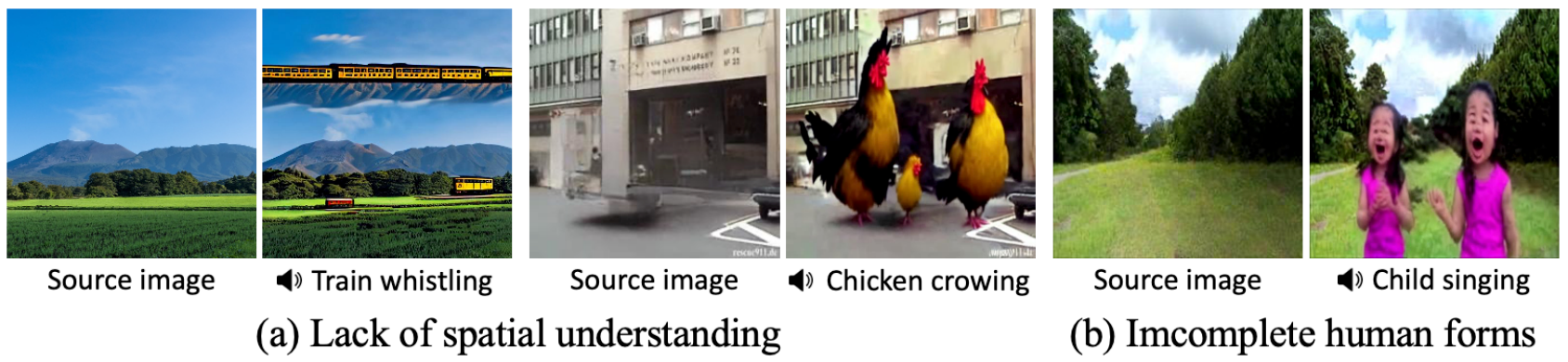}
    \vspace{-5mm}
    \caption{\textbf{Examples of failure cases.} We observe that our model occasionally inserts sounding objects into the scene without proper spatial understanding (a), or it tends to insert humans into the scene in incomplete forms (b).}
    \vspace{-2mm}
    \label{fig:limitation}
\end{figure}

\paragraph{Discussion}
Although our method shows promising results in editing with given audio inputs, we have observed limitations in failure cases. The first issue is that our model occasionally inserts objects without proper spatial understanding, as shown in \Fref{fig:limitation} (a). For example, the train may be inserted in the sky, or the chicken may appear as large as the building behind it. Another issue is that images edited to include human-related sound sources often exhibit low quality, as demonstrated in \Fref{fig:limitation} (b). This may be due to dataset bias, as our model is trained with video datasets featuring motion blur and relatively low resolution, which can result in unclear learning signals. Additionally, our model does not specify the location for inserting sounding objects but relies solely on the audio input. Incorporating text descriptions or learning from multichannel audio could address these issues, which we consider as a direction for future work.

\paragraph{Conclusion}
In this work, we introduce SoundBrush, a model that leverages audio signals to edit and manipulate visual scenes. By extending the capabilities of Latent Diffusion Models, we demonstrate the potential of using sound as a tool for image editing. Our approach involves constructing a sound-paired visual scene dataset, which is then used to train the model to translate diverse auditory cues into textual spaces, thereby enabling editing based on input sound. As a result, SoundBrush acquires the ability to finely manipulate scenes to reflect the mood of the input audio or to insert sounding objects while preserving the original structure. Moreover, we demonstrate that our model can be integrated with novel view synthesis methods, opening new possibilities for sound-guided 3D scene manipulation.

%% file: supp.tex
\renewcommand{\thefigure}{S\arabic{figure}}
\setcounter{figure}{0}

\noindent {\LARGE \textbf{Appendix}}\label{sec:appendix} \vspace{3mm}

In this supplementary material, we provide implementation details of our proposed SoundBrush and the baseline models, a detailed explanation of the dataset generation pipeline, additional analysis of 2D and 3D visual scene editing, and details about the user study experiments, which were not included in the main paper due to space limitations.

\input{supp_1.tex}
\input{supp_2.tex}

%% file: supp_1.tex
\section{Implementation Details}

\subsection{Training Details on SoundBrush}
Our proposed SoundBrush comprises a \textbf{mapping network} that translates audio features into the audio tokens, and an \textbf{image editing model} that uses these audio tokens to edit the given image. We train SoundBrush for 5,000 steps on 2x48GB NVIDIA RTX A6000 GPUs over 20 hours, with early stopping. Following the training scheme similar to InstructPix2Pix~\cite{pix2pix}, we train the model with the images at a resolution of 256x256 and a total batch size of 48. The image editing model, based on the Latent Diffusion Model (LDM), is initialized with a pretrained checkpoint from InstructPix2Pix, a text-based image editing model. Initially, our approach involves training only the mapping network to translate the input audio into the textual space of the image editing model, while keeping the image editing model frozen. Although the image editing model initialized with InstructPix2Pix had already learned a rich prior for editing visual scenes, we observe that it still struggles to insert new objects. Therefore, we apply Low-Rank Adaptation (LoRA) to the image editing model while jointly training with the mapping network.
We set the LoRA rank to 2 and the LoRA alpha value to 2. 
While our model is trained at a resolution of 256x256, we find that it generalizes well to larger resolution images during inference.
The results presented in this paper were generated at varying resolutions with 100 denoising steps, using an Euler ancestral sampler with the denoising variance schedule~\cite{karras2022elucidating}.

\subsection{Details on Baseline Models}\label{baseline}
We compare our proposed SoundBrush with existing models: InstructAny2Pix~\cite{any2pix}, AudioToken~\cite{audiotoken}, and GlueGen~\cite{gluegen}. The details of each model and their respective approaches for inference are as follows.

\paragraph{InstructAny2Pix}
InstructAny2Pix enables image editing with multimodal instructions including audio, images, and text. It employs a multimodal encoder, a diffusion model, and a multimodal LLM to interpret these instructions and edit images based on the source audio. The multimodal encoder extracts features from the input signals, and the multimodal LLM generates textual instructions from these features. These textual instructions are then fed into the diffusion model for visual scene editing.
We do not perform any fine-tuning, as we use a publicly available checkpoint trained on datasets, including VGGSound~\cite{vggsound}, which we also utilized for our experiments. 
For inference, we provide InstructAny2Pix with the instruction ``Add \{test audio\}'' to edit the specified visual scene.

\paragraph{GlueGen}
GlueGen is a model designed to generate images from scratch using input audio. It introduces the GlueNet model, which aligns the latent space of audio features with the text-to-image Stable Diffusion Model~\cite{ldm}. As GlueGen is primarily aimed at image generation rather than editing, we modify the self-attention layers (4th to 11th) and the 4th residual layer of the generation pipeline with a Plug-and-Play (PNP)~\cite{pnp} injection.
We set the guidance scale as 7.5 following the official repository.
Furthermore, the publicly available checkpoint was trained using the UrbanSound8K dataset~\cite{urban}, which differs from the dataset domain of our test samples. For a fair comparison, we fine-tune the model on the VGGSound dataset for 30 epochs using a learning rate of 2e-5. We follow the audio preprocessing and other hyperparameters summarized in their implementation details.

\paragraph{AudioToken}
AudioToken is a model specifically designed to generate images from scratch using sound. Similar to GlueGen, AudioToken leverages the existing text-to-image Stable Diffusion Model to accommodate sound as an input. Specifically, AudioToken utilizes a pre-trained audio encoder, BEATs~\cite{beats}, to extract audio features and designs a model that encodes these audio features into text tokens aligned with the CLIP text tokens.
Although primarily a generation model, we employ the same Plug-and-Play (PNP) approach as GlueGen to extend AudioToken's capabilities for image editing. 
We set the guidance scale at 7.5, following the official repository guidelines.
We do not perform any fine-tuning on AudioToken since the official model is trained with the VGGSound~\cite{vggsound} dataset, which is the same dataset we used for testing. We use the prompt ``A photo of a \{test audio\}'' to feed the audio into the Stable Diffusion Model for image editing.

%% file: supp_2.tex
\begin{figure}[tp]
    \centering
    \small
    \includegraphics[width=1.0\linewidth]{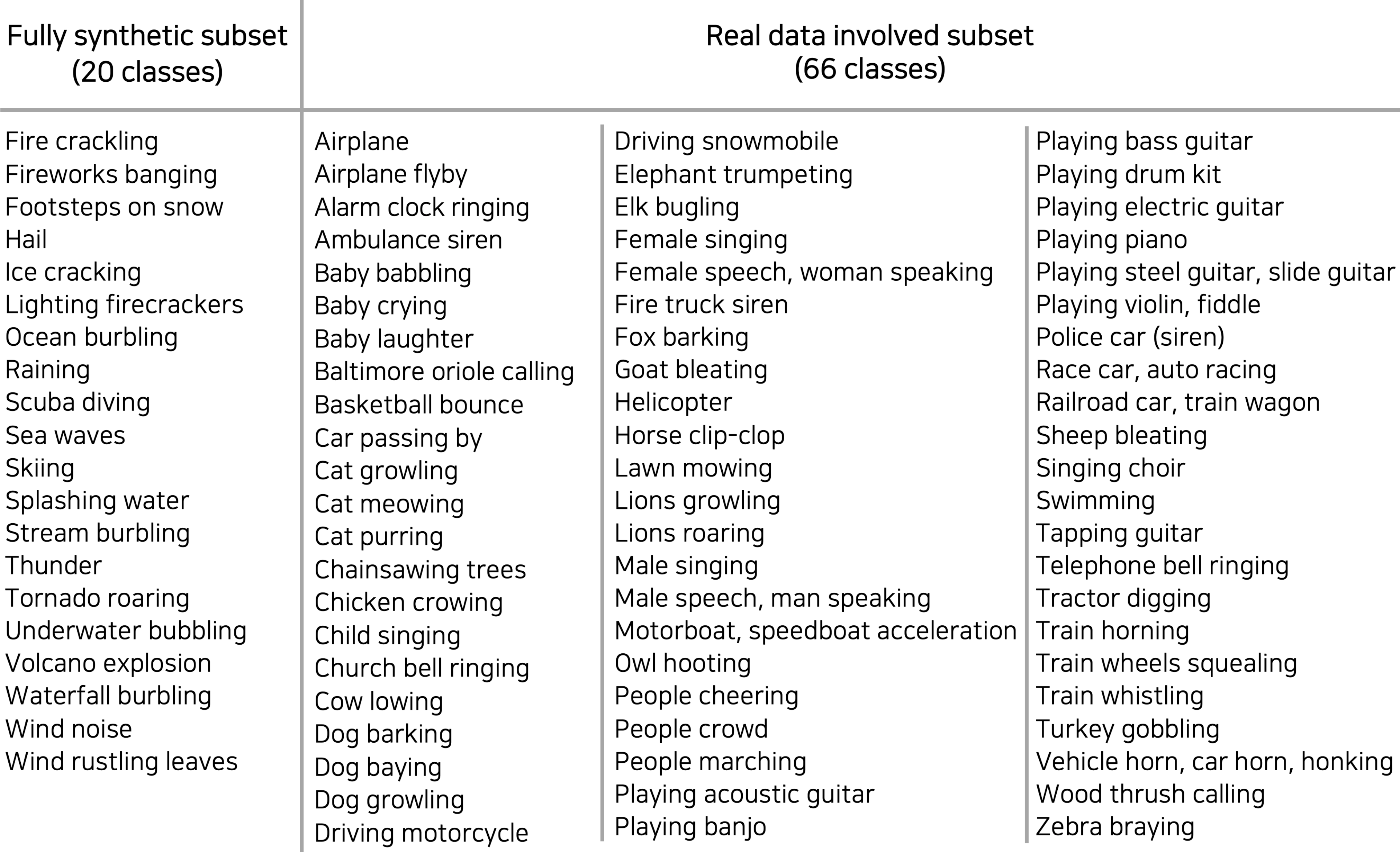}
    \caption{\textbf{Sound categories.} We select 86 different categories for constructing the dataset and training the SoundBrush model.}
    \label{supple:categories}
\end{figure}

\section{Dataset Construction}
We generate a paired dataset consisting of audio, and images before and after editing. As mentioned in the main paper, sound semantics vary significantly from one sound to another. For instance, a sound like ``Dog barking'' might only require the insertion of a dog into the scene, rather than altering the overall scene of the given image. In contrast, a sound like ``Footsteps on snow'' requires to manipulate the entire scene to accumulate snow on the ground, buildings, and trees. To accommodate these variations, we construct two types of subsets: a fully synthetic subset for learning the overall scene manipulation, and a real-data-involved subset for inserting sounding objects into the scene. The sound categories from the VGGSound dataset used to construct each subset are summarized in \Fref{supple:categories}.

\paragraph{Fully synthetic subset}
As described in the main text, generating the fully synthetic subset involves creating source and target prompts based on sound events and generating paired images from these prompts. 
Here, we illustrate the detailed process of each stage.

The process begins with generating source prompts that describe images of the general visual scenes. To generate diverse natural prompts, we utilize the generalizable generation capacity of the Large Language Model, GPT-4 mini~\cite{gpt4}, and provide instructions to produce diverse source prompts: 

\begin{boxA}
Example: ``An image of a city street on a sunny day'' 
\\
\\
Make N new diverse descriptions with a similar structure to the example above, but with different places and landmarks, and with diverse weather but mostly sunny.
\end{boxA}

Next, we generate the target prompt, incorporating audio information into source prompts. 
For this, we extract several sound categories from the VGGSound dataset~\cite{vggsound} as summarized in \Fref{supple:categories}.
However, simply adding the name of the sound category to the source prompt may not achieve the desired level of editing, as the sound category alone may not provide sufficient visual description. Therefore, we extract additional keywords for each sound category by analyzing the editing results on a small set of images. For example, the ``Raining'' category includes keywords such as ``Downpour,'' ``Heavy rain,'' and ``Rain shower.'' These audio-related keywords are then used to generate source-target pairs. We again utilize GPT-4 mini to create diverse and natural prompt pairs, guiding it with exemplary pairs as shown below:

\begin{boxA}
- Source : The marketplace had fruit stands under the street lamp against the sunny afternoon sky.

- Keyword : thunder

- Target : The marketplace had fruit stands and a street lamp, with lightning in the sky.
\\
\\
By referencing the above examples, write sets of ``target'' for each ``label'' based on given ``source''. Print the label, and target in order, but do not print the source.
\\
\\
- Source: $\textit{Source prompt}$

- Keyword : $\textit{Audio Keywords}$

- Target : /// Fill in ///
\end{boxA}

Finally, we create 10,000 source and target prompt pairs using the aforementioned process and apply Prompt2Prompt~\cite{p2p} to generate image pairs from these prompts. In Prompt2Prompt, the p-value controls the degree of editing. However, our experiments show no significant variation in image editing with different p-values, thus we fix the p-value at 0.5. Instead, we use five different initial noises for each pair generation.

Due to the stochastic nature of generation methods, not all results are of high quality. Therefore, as detailed in the main paper, we employ several CLIP-based and ImageBind-based metrics to filter out noisy pairs. We set the thresholds as follows: 0.2 for directional similarity in CLIP space, 0.7 for feature similarity between the source and target images, and 0.2 for audio-visual feature similarity between the edited image and the sound in ImageBind space. Samples falling below these thresholds are discarded.

\begin{figure*}[tp]
    \centering
    \small
    \includegraphics[width=1.0\linewidth]{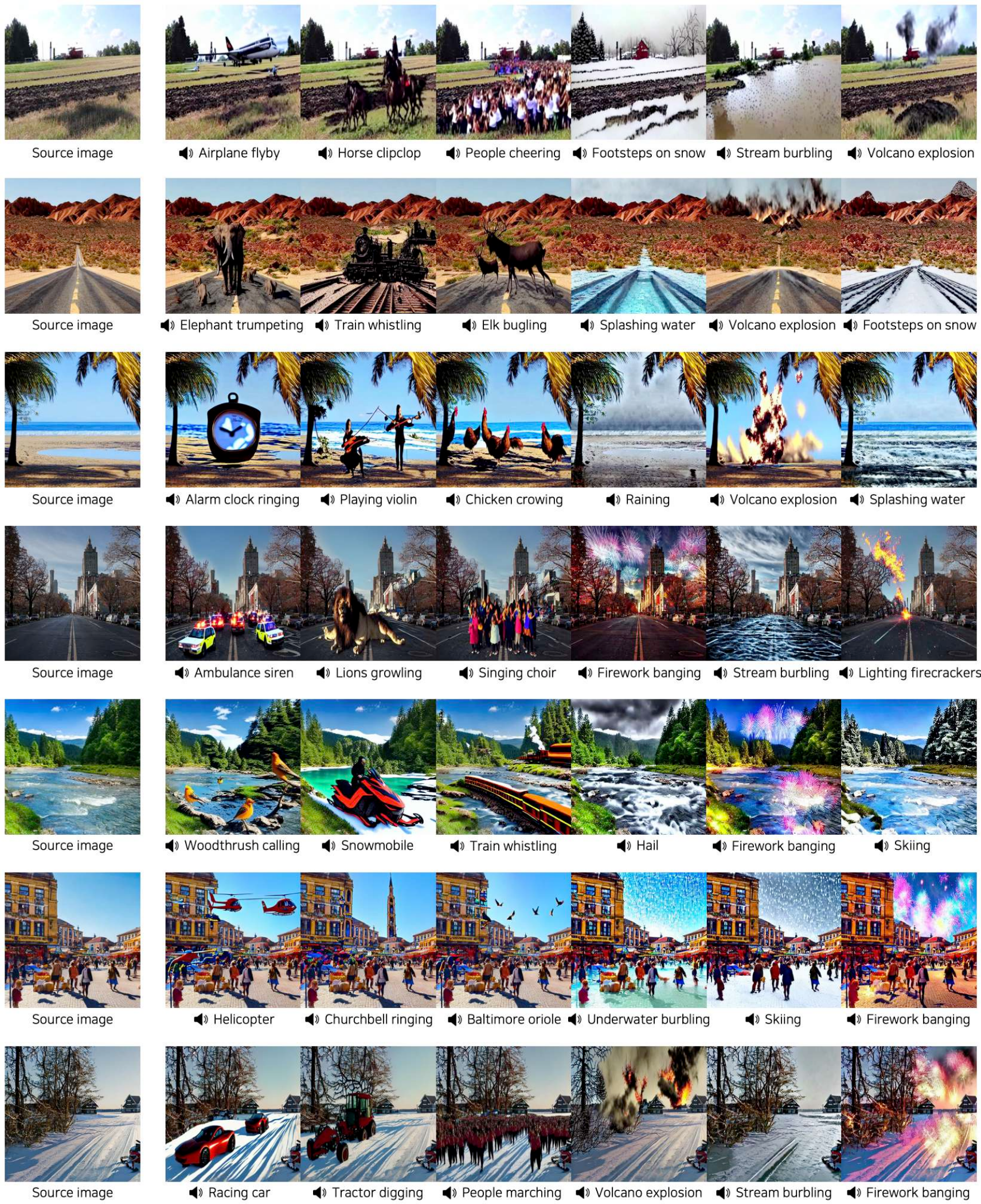}
    \caption{\textbf{Additional qualitative results of our SoundBrush.} From the same source image, SoundBrush can edit the given image using a diverse range of in-the-wild sounds, including environmental sounds, animals, vehicles, and human-produced sounds.}
    \label{supple:qual}
\end{figure*}

\begin{figure}[tp]
    \centering
    \small
    \includegraphics[width=1.0\linewidth]{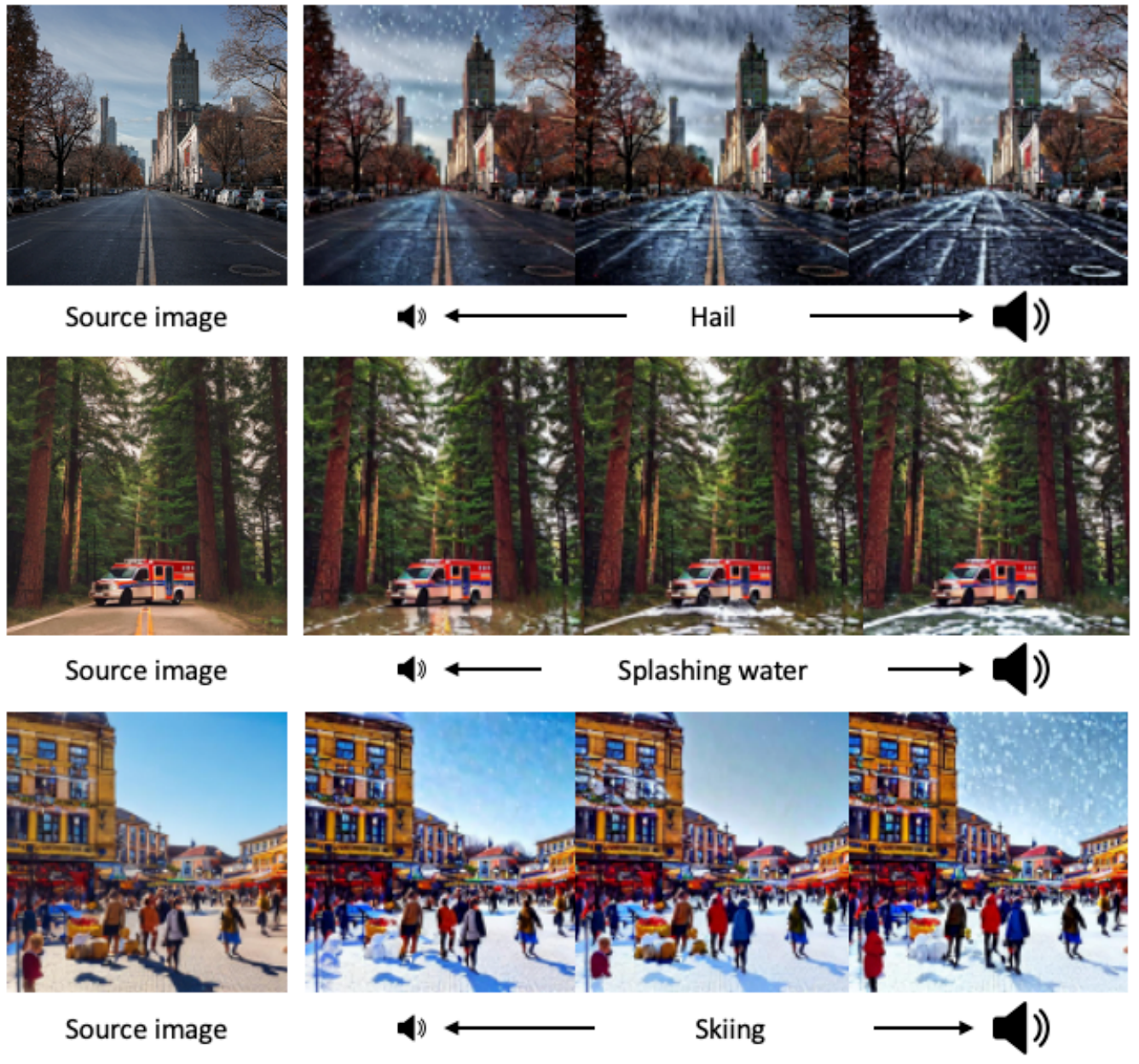}
    \caption{\textbf{Edited visual scene by varying the volumes of the input audio.} SoundBrush captures the intensity differences in the audio and reflects these changes in the edited images.}
    \label{supple:volume}
\end{figure}

\begin{figure}[t]
    \small
    \centering
    \includegraphics[width=0.95\linewidth]{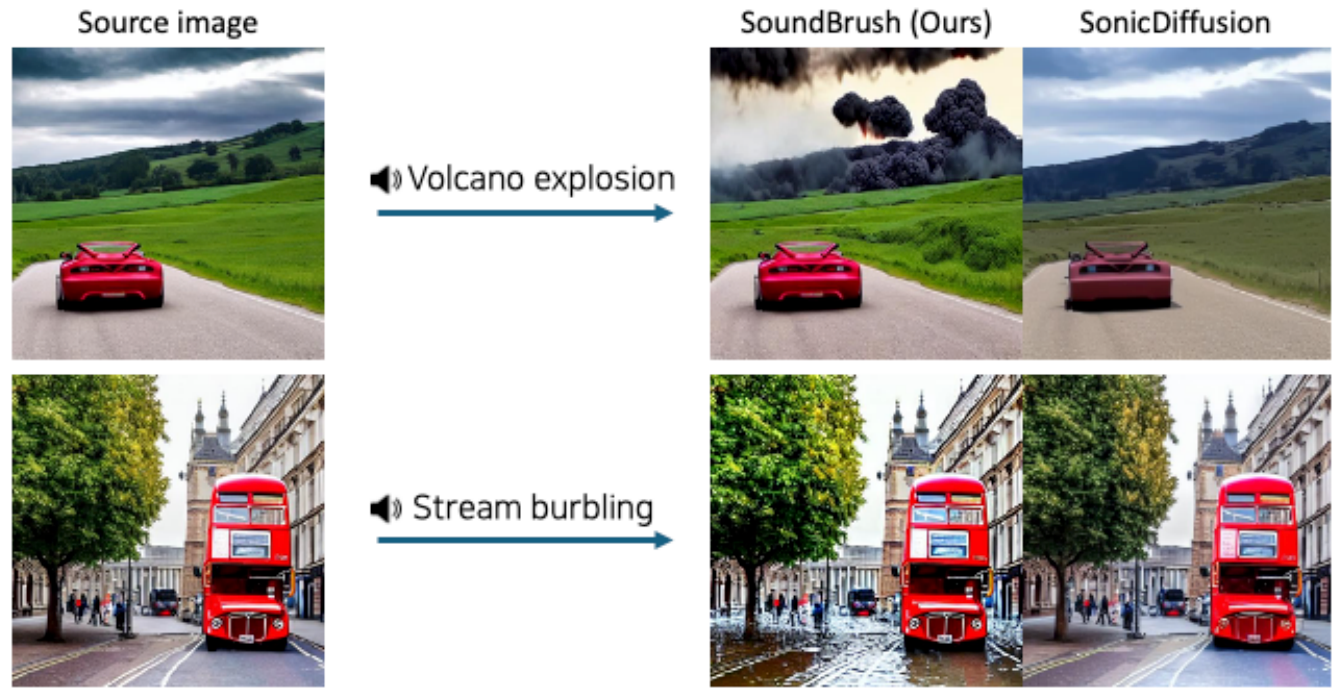} \\
    \resizebox{0.95\linewidth}{!}{
    \begin{tabular}{lcccc}
    \toprule
     Method & AVS ($\uparrow$) & IIS ($\uparrow$) & TVS ($\uparrow$) & FID ($\downarrow$)\\
    \cmidrule{1-5} 	
    SonicDiffusion & 0.158& 0.742& 0.155 & \textbf{129.5}\\
    SoundBrush (Ours) & \textbf{0.261}& \textbf{0.772}& \textbf{0.203} & 131.7\\
    \bottomrule\\
    \end{tabular}
    }
    
    \caption{\textbf{Comparison.} We qualitatively and quantitatively compare SoundBrush with the recent image generation model, SonicDiffusion~\cite{sonicdiffusion}.}
    \label{supple:quan}
    \vspace{-1mm}
\end{figure}

\paragraph{Real data involved subset}
As described in the main paper, we use a real image and audio from the VGGSound dataset~\cite{vggsound} and apply a sound source localization model to identify and segment the sounding object in the image. We then employ an inpainting model, which inpaints the masked area of the given image. This process allows us to construct the subset that contains the original real image as the ``after editing'' image and the inpainted image as the ``before editing'' image in the subset.

Sound source localization is a technique that identifies the sound source within a visual scene based on the input audio. Among the models, we use \citet{park2024can}, which has demonstrated impressive results by utilizing CLIPSeg~\cite{luddecke2022image} as the localization module. For the inpainting model, we employ LaMa~\cite{suvorov2022resolution}, which takes an image and corresponding mask as inputs and effectively inpaints the masked areas. LaMa has shown a remarkable ability to adapt to diverse scenes, ranging from low to high-resolution images. Despite the varying quality of the real images in our video dataset, LaMa consistently delivers robust inpainting results when the masks are appropriately provided.

As noted in the main paper, this automated pipeline is not always perfect and often produces noisy sets. To further filter out these noisy sets, we utilize several off-the-shelf models to establish criteria for filtering. We measure the image-image similarity between the before and after inpainting images using CLIP~\cite{clip}, setting a threshold of 0.7. If the similarity exceeds this value, the sample is considered not appropriately inpainted and is thus discarded. Additionally, we use ImageBind~\cite{girdhar2023imagebind} as another filtering criterion. If the feature similarity between the audio and the inpainted image exceeds that between the audio and the before-inpainted image, we determine that the inpainted image still contains the sounding object and discard it.

\begin{figure*}[tp]
    \centering
    \small
    \includegraphics[width=1.0\linewidth]{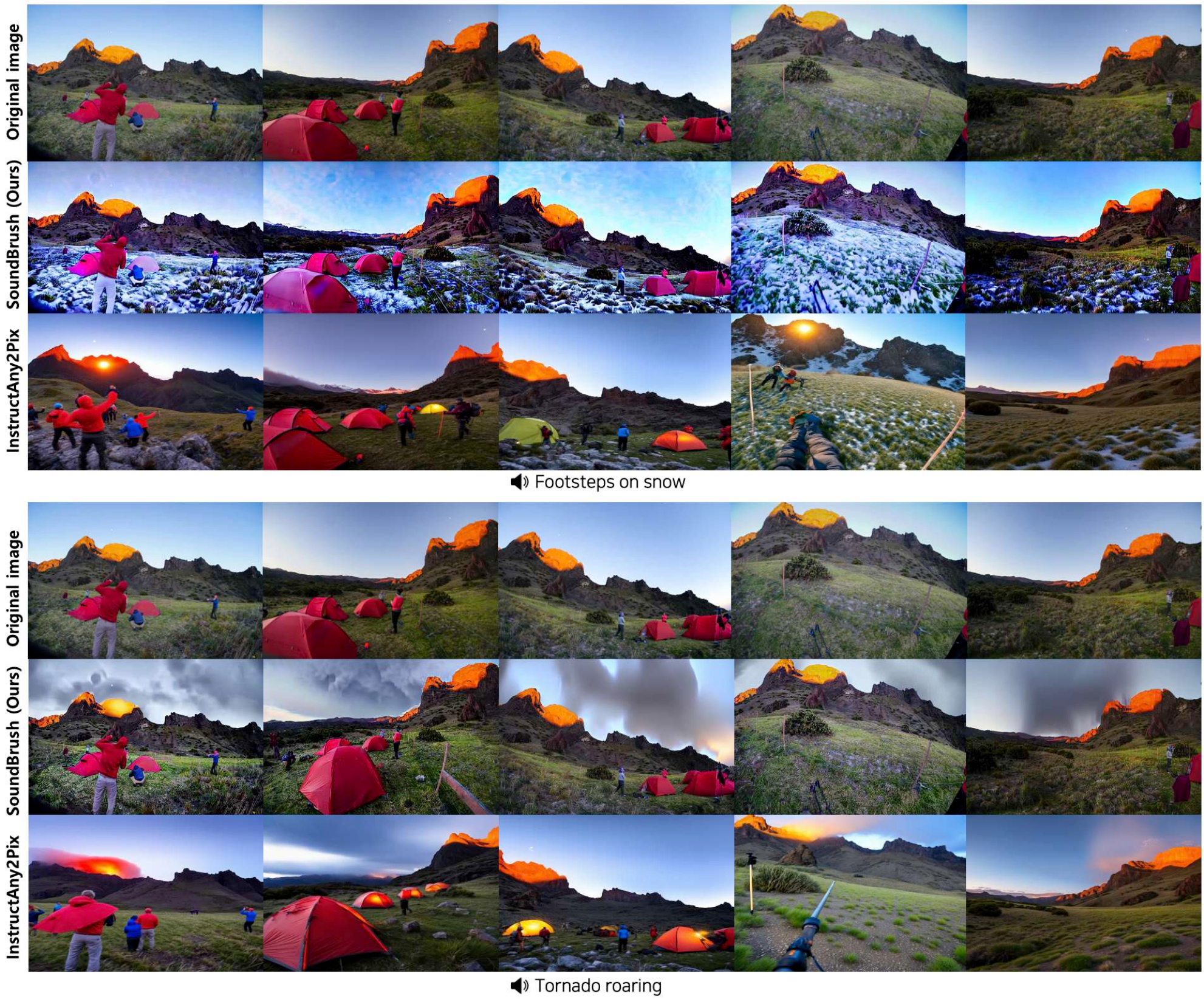}
    \caption{\textbf{Visualization of intermediate 2D edited images during 3D visual scene editing.} InstructAny2Pix alters the original content during editing. In contrast, SoundBrush effectively manipulates the visual scene to reflect the sound semantics while preserving the original structure, thus stabilizing the 3D visual scene editing process.}
    \label{supple:3D}
    
\end{figure*}

\begin{figure*}[tp]
    \centering
    \small
    \includegraphics[width=1.0\linewidth]{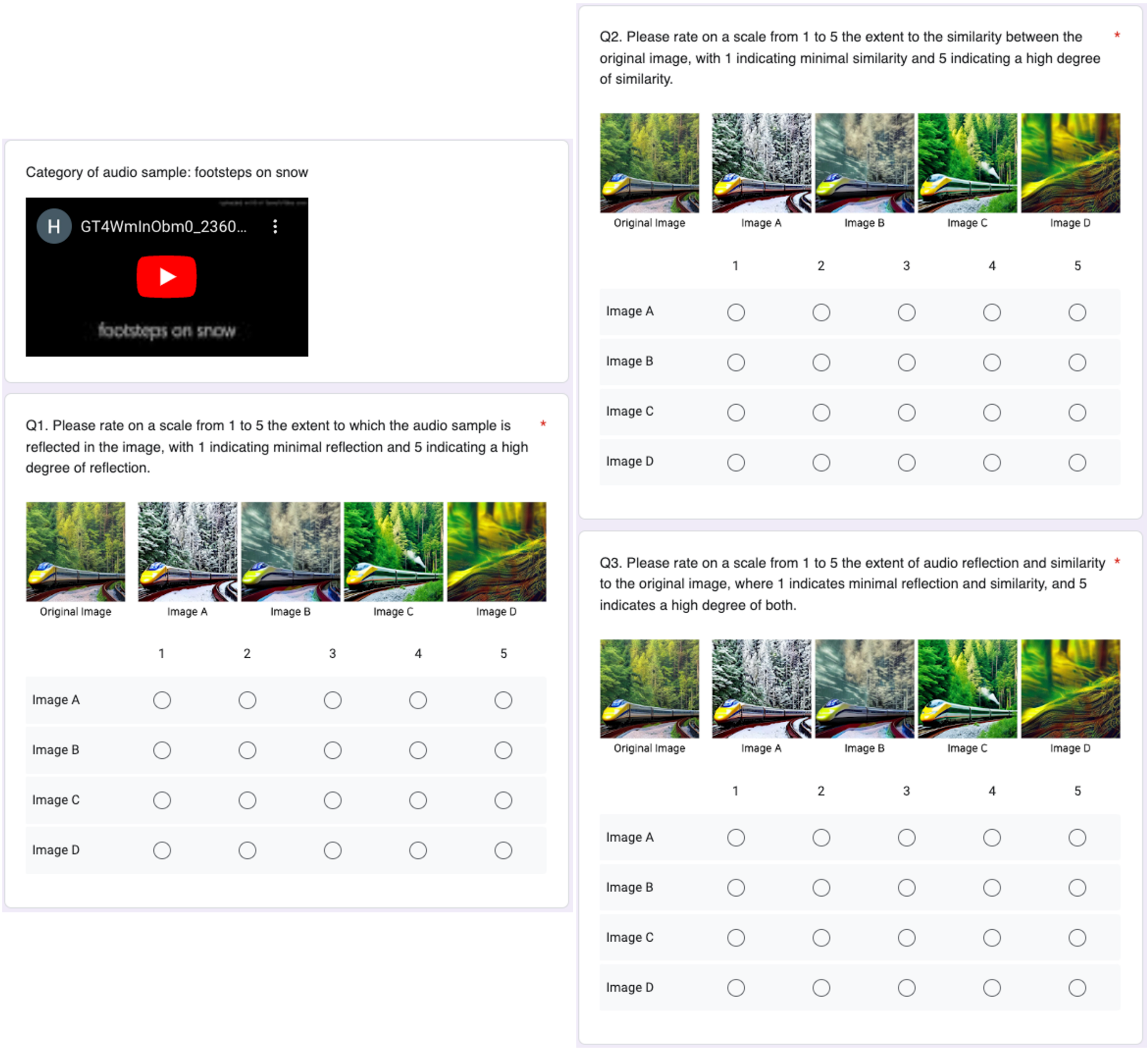}
    \caption{\textbf{Examples of the user study.} We conduct a user study to compare our method with InstructAny2Pix, AudioToken, and GlueGen. The goal of the user study is to evaluate the edited images from the following perspectives: Q1. How well the input audio is reflected in the image, Q2. How well the original structure and content are preserved, and Q3. How effectively the audio semantics are reflected while keeping the original structure unchanged. We assess the performance of each model using a Mean Opinion Score (MOS) test, with ratings defined as follows: 1 = Bad, 2 = Poor, 3 = Fair, 4 = Good, and 5 = Excellent.}
    \label{supple:human}
\end{figure*}

\section{Additional Analysis}
\subsection{Results on 2D Editing}
We present additional results for 2D visual scene editing in \Fref{supple:qual}. Our model effectively handles a diverse range of input source images and uses in-the-wild sound as a conditioning signal to tailor edits to the visual scene, accurately reflecting the input sound. For example, given a source image of a city with buildings, as shown in the fourth row, we can insert various sounding objects, such as vehicles, animals, and groups of people onto the road. We can also manipulate the overall scene to include semantics from diverse environmental sounds, such as fireworks, water, or lighting firecrackers.

\subsection{Experiments on Volume Changes}
Unlike textual signals, sound carries unique information related to intensity. We provide additional examples to demonstrate how our model captures these intensity changes to manipulate the visual scene effectively. As illustrated in \Fref{supple:volume}, the effects on the edited image become more dynamic in response to audio inputs with varied intensities. For example, as the ``Hail'' sound intensifies, the raindrops appear harder; similarly, as the ``Splashing water'' sound increases, the water splashing below the vehicle becomes more dynamic.

\subsection{Comparison with SonicDiffusion}
We also compare our SoundBrush with SonicDiffusion~\cite{sonicdiffusion}. Similar to GlueGen and AudioToken, SonicDiffusion is built on the LDM to generate images from audio inputs. As SonicDiffusion is primarily an image generation model, the PNP method is adopted for editing tasks. As shown in \Fref{supple:quan}, our SoundBrush outperforms SonicDiffusion both quantitatively and qualitatively. Specifically, our model excels in editing-related metrics while showing comparable FID scores. We observe that SonicDiffusion often makes minimal changes to the original image, thus failing to edit the image in alignment with the input audio.

\subsection{Analysis on 3D Editing}
InstructNeRF2NeRF~\cite{n2n} is trained by progressively replacing input training images with edited ones, thus facilitating the progressive editing of 3D scenes. We evaluate how our proposed SoundBrush and InstructAny2Pix edit images that are subsequently used to replace the training images in InstructNeRF2NeRF. We sample edited images from the first 30 iterations and visualized them in \Fref{supple:3D}. Although InstructAny2Pix~\cite{any2pix} incorporates sound semantics into the edited images, it often alters the original scene placement or content, leading to unstable 3D editing. In contrast, SoundBrush effectively integrates sound input while preserving the original content, thereby enhancing the stability of 3D editing compared to InstructAny2Pix.

\section{Details of the User Study}
As the image editing task is challenging to evaluate solely through quantitative metrics, we conduct a user study to analyze the performance of our method from a human perception perspective. The user study questionnaire interface, shown in \Fref{supple:human}, allows users to listen to the given audio, view the source image, and then assess the edited images from four different methods without any time constraints. The methods include our proposed SoundBrush, InstructAny2Pix~\cite{any2pix}, AudioToken~\cite{audiotoken}, and GlueGen~\cite{gluegen}, presented in random order. The user study comprised 20 samples, each with three questions, totaling 60 questions.

Although the primary objective of image editing is to reflect the given condition while preserving the original structure of the image, we design separate questions to evaluate each component of the image editing task. We assess the performance of each model using a Mean Opinion Score (MOS) test conducted with 47 subjects. Users are asked to rate on a scale from 1 to 5 based on the following questions: Q1. How well the input audio is reflected in the image, Q2. How well the original structure and content are preserved, and Q3. How effectively the audio semantics are reflected while keeping the original structure unchanged. The ratings for each answer are defined as: 1 = Bad (unacceptable quality), 2 = Poor (barely acceptable quality), 3 = Fair (acceptable but with some dissatisfaction), 4 = Good (satisfactory quality), and 5 = Excellent (high quality and fully satisfactory).

The objective of Q1 is to determine whether the audio semantics are well-reflected in the given image, regardless of how much the original content has been altered. Therefore, if a model neglects the original content and generates the image solely from the given sound, it may still receive a high rating.
The objective of Q2 is to assess whether the original structure and content are preserved, regardless of whether the audio semantics are reflected. Consequently, if a model fails to edit and outputs an image identical to the source, it may also receive a high rating.
Finally, Q3 combines Q1 and Q2 to evaluate how audio semantics are reflected while keeping the original structure unchanged, which is the primary goal of this user study. Interestingly, as reported in the main paper, our method outperforms the existing methods in every questionnaire, demonstrating the effectiveness of our approach for sound-guided image editing.